\documentclass{article}

\usepackage{arxiv_default}
\usepackage{amsmath,amssymb,booktabs,graphicx,microtype,multirow,xcolor}
\usepackage{hyperref}
\usepackage{cleveref}

\newcommand{\ours}{\textsc{Self-Generated T2T}}
\newcommand{\oursshort}{\textsc{Ours}}

\newcommand{\mask}{\texttt{[MASK]}}
\newcommand{\casebox}[1]{%
  \par\medskip\noindent
  \fbox{\parbox{\dimexpr\linewidth-2\fboxsep-2\fboxrule}{#1}}%
  \par\medskip}

\title{Self-Generated Error Training for Token Editing in Diffusion Language Models}

\author{
  Lin Yao$^{1,2}$ \\
  $^1$School of Computer Science, Shanghai Jiao Tong University, Shanghai, 200240, China \\
  $^2$Zhongguancun Academy, Beijing, 100097, China \\
  \texttt{lin.yao@sjtu.edu.cn}
}

\begin{document}

\maketitle

\begin{abstract}
Token-to-token (T2T) editing lets LLaDA2.1 revise committed tokens during block-diffusion decoding.
The released recipe trains this editor on random vocabulary corruptions, but at inference the editor sees the model's own fluent, high-confidence draft errors instead.
We study this training-inference mismatch and propose \ours{}, which performs a no-gradient draft pass, fills masked positions with predicted tokens, and supervises recovery in a second pass under these self-generated corruptions.
We implement the update as a short LoRA continued-pretraining pass on LLaDA2.1-mini and evaluate on several benchmarks under the official Q-Mode T2T procedure with unchanged inference parameters.
The method generally improves accuracy while reducing T2T edit intensity, mitigating failure modes such as final-digit transcription errors after otherwise correct reasoning and excessive self-correction before short factual answers.
\end{abstract}

\section{Introduction}
\label{sec:intro}

Diffusion language models generate text by denoising partially observed sequences rather than committing to a strictly left-to-right trajectory.
This gives them two attractive properties: they can update many positions in parallel, and they can revisit previous commitments.
LLaDA2.1 operationalizes the second property through token-to-token (T2T) editing~\citep{llada21}.
Each block is decoded with standard mask-to-token (M2T) filling, followed by an editing phase in which visible tokens are re-scored and overwritten when a new candidate exceeds an editing threshold.
This edit path is central to LLaDA2.1's quality-speed tradeoff: it allows the model to accept tokens earlier and repair them later.
It is also necessary because parallel denoising can commit multiple tokens under the same incomplete context: tokens predicted in the same step cannot condition on one another while the corresponding positions are still masked, so locally plausible commitments may become inconsistent after they are placed together.

The training recipe behind this editor is less aligned with deployment.
During training, the T2T stream is constructed by selecting clean tokens, replacing them with random vocabulary tokens, and applying cross-entropy to recover the original tokens.
During inference, the tokens that need editing are not random samples from the vocabulary.
They are the model's own earlier predictions: plausible numbers in a derivation, locally grammatical but wrong entities, premature answer tokens, or semantically related alternatives.
These errors are harder than random corruptions because they are in-distribution as language and can actively steer the context.

This paper asks whether the T2T editor should be trained on the errors it will actually see.
We propose \ours{}, a training-side alignment method for LLaDA2.1 T2T editing.
Instead of generating visible corruptions entirely from a uniform vocabulary distribution, we first ask the current model to fill masked positions, then use those predicted tokens as visible inputs for a second supervised pass: incorrect predictions are trained as edits back to the clean tokens, while correct predictions are trained to remain unchanged.
The training objective remains a supervised denoising objective; inference keeps the original LLaDA2.1 T2T procedure and official Q-Mode parameters unchanged.
The only change is the source of T2T corruption.

\begin{figure}[htp]
\centering
\includegraphics[width=\linewidth]{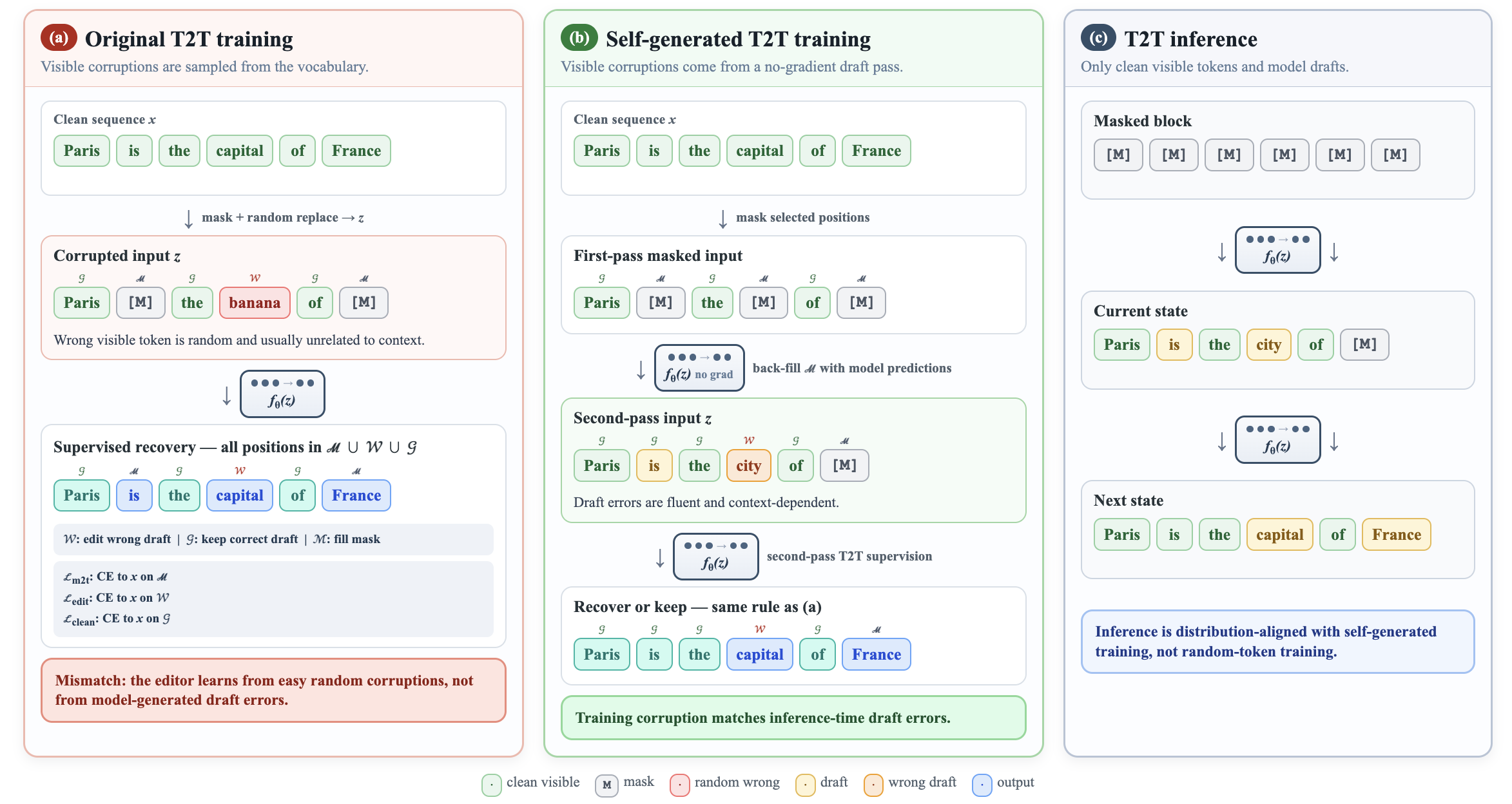}
\caption{\textbf{Training-inference alignment for T2T editing.}
\textbf{(a)} Random-token training corrupts visible tokens with unrelated vocabulary (e.g., \emph{capital}$\to$\emph{banana}), yielding off-manifold errors.
\textbf{(b)} \ours{} drafts masked positions with self-generated tokens and trains on context-dependent errors from the same distribution.
\textbf{(c)} Inference edits the same on-manifold drafts; \ours{} CPT aligns training with deployment without changing decoding.}
\label{fig:method_overview}
\end{figure}

We evaluate \ours{} on CMATH, TriviaQA, PIQA, and AIME~2025 under the official Q-Mode T2T inference protocol (\cref{sec:data_model}), changing only the checkpoint weights while keeping inference procedure and parameters fixed.
Table~\ref{tab:main_results} reports accuracy and generation traces on these four benchmarks.
On CMATH, TriviaQA, and PIQA, \oursshort{} improves accuracy while reducing T2T edit intensity; on AIME~2025 accuracy is unchanged at 9/30, but edits still drop from 130.2 to 86.0 per problem.
CMATH shows the largest gain (+5.47 accuracy points) even though average output length increases, consistent with better final numeric commitment rather than premature short answering.
This is a deliberately minimal LoRA CPT study on a convenience FineWeb subset (\cref{sec:scope}), intended as a mechanism stress test rather than a fully optimized recipe.

Our contributions are:
\begin{itemize}
    \item We formulate a training-inference mismatch in LLaDA2.1 T2T editing: random-token training corruptions do not match structured model-generated inference errors.
    \item We propose \ours{}, a two-pass continued-training objective that trains the native T2T editor on self-generated draft tokens without adding a detector, correction head, RL stage, or inference-time mechanism.
    \item We provide a reproducible LoRA implementation on LLaDA2.1-mini with a convenience FineWeb-Edu CPT setup, including corruption parameters, loss weights, optimizer settings, and evaluation under the official Q-Mode T2T inference defaults.
    \item We report results on four benchmarks under the same official inference procedure and parameters: CMATH (+5.47 accuracy points), TriviaQA, and PIQA improve with lower edit intensity, while AIME~2025 matches the base score but still needs fewer T2T edits.
\end{itemize}

\section{Related Work}
\label{sec:related}

\paragraph{Discrete diffusion LMs and parallel commitment.}
Discrete diffusion models extend denoising diffusion to categorical state spaces, with absorbing-mask corruption providing a natural bridge to masked language modeling~\citep{austin2021d3pm}.
Language-specific variants refine this view into practical text-generation objectives and architectures, including SEDD, MDLM, MD4, and RADD~\citep{lou2024sedd,sahoo2024mdlm,shi2024md4,ou2025radd}.
Recent systems scale this family in different but related ways.
Some train large masked diffusion LMs directly, some initialize from autoregressive checkpoints and continue training them as diffusion LMs, and some change the decoding architecture through blockwise diffusion for higher throughput; representative examples include LLaDA, Dream, DiffuLLaMA, Mercury, BD3-LM, and LLaDA2.1~\citep{nie2025llada,dream7b,gong2025diffullama,inception2025mercury,arriola2025bd3lm,llada21}.
The central advantage of these models is parallel denoising: multiple positions can be predicted before a strict left-to-right context is available, and recent analyses study the resulting tradeoff between speed, generation order, and quality~\citep{kang2025parallelbench}.
The same advantage creates the revision problem studied in this paper.
Tokens predicted in the same parallel step share the same partially observed context, but they cannot condition on the other tokens being predicted in that step because those positions are still masked at prediction time.
As a result, the model can commit tokens that are individually plausible under the shared context but mutually inconsistent after they are placed together; they may disagree semantically, duplicate incompatible roles, or form locally plausible but globally invalid spans.
Once committed, these independently predicted tokens become visible context for later denoising steps, so a local conflict can propagate into subsequent predictions.
Thus, high-quality parallel denoising requires some mechanism for revising earlier commitments.

\paragraph{Editing and remasking mechanisms.}
Existing revision mechanisms differ mainly in the action they take on a suspicious visible token.
LLaDA2.1 introduces token-to-token (T2T) editing: after standard mask-to-token (M2T) denoising, the model may overwrite an already visible token when a new candidate becomes sufficiently confident~\citep{llada21}.
Remasking methods instead undo the commitment by resetting uncertain visible tokens to \mask{} and letting the M2T stream re-predict them; examples include inference-time remasking samplers, self-reflective remasking, and token-to-mask refinement~\citep{wang2025remdm,huang2026remedi,remask_dont_replace}.
Learned self-correction methods add still another axis: ProSeCo inserts explicit corrective refinement steps, while BackPlay trains a lightweight corrector on errors made by a frozen generator~\citep{schiff2026proseco,liu2026backplay}.
These mechanisms address the same consequence of parallel commitment, but they intervene at different points: changing the inference action, adding a correction module, or adding a correction phase.
They are complementary to our work and could in principle be combined with it: a model trained on realistic editable-token distributions can still use replacement, remasking, or an additional correction module at inference.
To isolate the training-distribution effect, we deliberately use the simplest correction action in this paper.
We keep the native LLaDA2.1 token-to-token replacement action and inference algorithm fixed, and change only the training distribution of the native editor.

\paragraph{The T2T corruption mismatch.}
The T2T editor in LLaDA2.1 is trained to recover clean tokens from randomly substituted visible tokens~\citep{llada21}.
This is a convenient supervised objective, but it does not match the editable tokens encountered at inference.
At inference, the tokens that should be edited are themselves selected by earlier model predictions.
Because the editor was trained on random replacements, its estimate of which visible tokens are worth editing, and what they should be edited into, can be poorly calibrated for fluent model-generated errors.
Such errors are qualitatively different from uniformly sampled vocabulary tokens: they are context-dependent, semantically plausible, and can steer the surrounding context.
Recent remasking work notes the same mismatch and avoids replacement by moving suspicious tokens back to \mask{}~\citep{remask_dont_replace}.
Our response is orthogonal to the choice between replacement and remasking.
Whether a system ultimately edits a token or remasks it, the training distribution should expose the model to realistic model-generated errors; in this work we keep the native LLaDA2.1 replacement action and make its T2T training distribution closer to deployment.

\paragraph{Training on model-induced states.}
A recurring way to reduce train-test mismatch is to train on states induced by the model itself, but the relevant state depends on the generation paradigm.
In autoregressive sequence modeling, scheduled sampling replaces some gold prefix tokens with samples from the model, exposing the next-token predictor to model-generated prefixes rather than only teacher-forced prefixes~\citep{bengio2015scheduled}.
In imitation learning, DAgger rolls out the current policy, queries the expert on the states that policy actually visits, and aggregates those policy-induced states into the training set~\citep{ross2011dagger}.
In diffusion models, self-conditioning uses the model's own intermediate denoising prediction as an additional input for subsequent denoising, so training better matches the iterative structure used at sampling time~\citep{chen2023analogbits}.
These methods share a principle, but not a state space: autoregressive models induce prefixes, imitation policies induce sequential decision states, and diffusion denoisers induce intermediate estimates of the clean data.
For T2T editing in discrete diffusion LMs, the induced state is different again: a partially denoised token block containing visible draft tokens produced by earlier denoising steps, some of which are plausible but wrong.
LLaDA2.1 trains its editor on random token substitutions, whereas inference asks the editor to repair model-generated visible token errors.
\ours{} instantiates self-generated T2T training for this setting by using the model itself to create the corrupted visible tokens used for T2T supervision.

\section{Method}
\label{sec:method}

\subsection{Base LLaDA2.1 Training and Inference}
\label{sec:background}

\paragraph{Base LLaDA2.1 training.}
For training, let $\mathbf{x}=(x_1,\ldots,x_n)$ be a clean sequence, where $i\in\{1,\ldots,n\}$ indexes token positions.
A corrupted training input $\mathbf{z}$ contains three kinds of positions.
Positions in $\mathcal{M}$ are mask-input positions filled with \mask{}.
Positions in $\mathcal{W}$ are visible but wrong-token positions.
Positions in $\mathcal{G}$ are visible ground-truth-token positions.
The model reads the full corrupted input $\mathbf{z}$, including all mask, wrong-token, and ground-truth-token positions, and produces one token distribution per position:
\begin{equation}
    (p_1,\ldots,p_n)=f_\theta(\mathbf{z}).
\end{equation}
For all supervised positions, the target is the clean token $x_i$.
The three position types therefore correspond to
\begin{equation}
\label{eq:three_losses}
    \mathcal{L}_{\mathrm{m2t}}=\sum_{i\in\mathcal{M}}\mathrm{CE}(p_i,x_i),\quad
    \mathcal{L}_{\mathrm{edit}}=\sum_{i\in\mathcal{W}}\mathrm{CE}(p_i,x_i),\quad
    \mathcal{L}_{\mathrm{clean}}=\sum_{i\in\mathcal{G}}\mathrm{CE}(p_i,x_i).
\end{equation}

Starting from $\mathbf{x}$, LLaDA2.1 training constructs $\mathbf{z}$ by first sampling a mask ratio and replacing the corresponding positions with \mask{}.
Among the remaining visible positions, a subset is then selected for token-level corruption and replaced by random vocabulary tokens different from the clean tokens.
The uncorrupted visible positions keep their ground-truth tokens.
M2T trains on the mask-input positions: given the full $\mathbf{z}$ as context, the model predicts the clean tokens at positions in $\mathcal{M}$, corresponding to $\mathcal{L}_{\mathrm{m2t}}$.
T2T trains on the visible positions: wrong visible tokens in $\mathcal{W}$ should be edited back to the clean tokens, and already-correct visible tokens in $\mathcal{G}$ should be kept unchanged, corresponding to $\mathcal{L}_{\mathrm{edit}}+\mathcal{L}_{\mathrm{clean}}$.

\paragraph{LLaDA2.1 T2T inference.}
At inference, the state $\mathbf{z}^t$ again contains mask-input and visible-input positions, but the visible tokens are produced by earlier model denoising steps rather than by random replacement.
M2T fills selected mask-input positions according to confidence thresholds.
T2T evaluates visible-input positions and replaces a current visible token when the model's top candidate at that position exceeds the editing threshold and differs from the current token.
Thus, base T2T exposes the editor to random-token visible corruptions during training, while inference-time T2T edits visible tokens generated by the model itself.

\subsection{Self-Generated T2T}
\label{sec:self_generated}

\ours{} keeps the target sequence, model architecture, and inference algorithm unchanged.
It changes only how the wrong visible tokens in $\mathcal{W}$ are constructed during training.
For each batch, we first run the current model without gradients on a masked input.
We then select a subset of masked positions and back-fill them with self-generated draft tokens from a no-gradient argmax pass, which may be correct or incorrect.
Incorrect drafts become wrong visible tokens in $\mathcal{W}$, while correct drafts become clean visible tokens in $\mathcal{G}$.
With probability $\rho_{\mathrm{rand}}=0.05$, a back-filled position is replaced by a random vocabulary token instead of the self-generated draft.
The second pass uses two forward passes per example (\texttt{num\_forwards=2}) on the resulting draft.
The second-pass input $\mathbf{z}$ has the same structure as above: \mask{} tokens on $\mathcal{M}$, self-generated or random wrong tokens on $\mathcal{W}$, and ground-truth tokens on $\mathcal{G}$.
Training uses the same three losses in \cref{eq:three_losses} with $(\lambda_{\mathrm{m2t}},\lambda_{\mathrm{edit}},\lambda_{\mathrm{clean}})=(1.0,0.3,0.2)$:
\begin{equation}
    \mathcal{L}
    =
    \lambda_{\mathrm{m2t}}\mathcal{L}_{\mathrm{m2t}}
    +\lambda_{\mathrm{edit}}\mathcal{L}_{\mathrm{edit}}
    +\lambda_{\mathrm{clean}}\mathcal{L}_{\mathrm{clean}}.
\end{equation}
In contrast to random-token T2T, the wrong-token positions $\mathcal{W}$ now mostly contain errors produced by the model's own first-pass denoising, which is closer to the visible-token errors encountered by T2T inference.

\subsection{Compute-Constrained Study Design}
\label{sec:scope}

The principled way to address the T2T training-inference mismatch is to use model-generated visible corruptions whenever the model is trained with a T2T objective.
Ideally, this would start from scratch: during pretraining, supervised instruction tuning, and any later alignment stage, the editor would learn to repair errors induced by the current model rather than random vocabulary substitutions.
Such full-scale training would be the cleanest validation of our hypothesis, but it is beyond our available compute.

We therefore use a deliberately minimal validation.
Starting from the released LLaDA2.1-mini checkpoint, which has already gone through pretraining, downstream instruction tuning, and alignment, we add only a low-rank LoRA update and run a short continued-pretraining experiment on English educational prose from FineWeb-Edu.
This setting is not intended as the final recipe.
It is a stress test of the mechanism: if replacing random T2T corruptions with self-generated corruptions is useful, the signal should be visible even when only a small adapter is trained on top of an already-finished model with out-of-domain CPT text.
The results below should therefore be read as evidence that the training distribution matters for T2T editing, not as a claim that the current LoRA checkpoint is the best possible model.

\subsection{Training and Inference Setup}
\label{sec:data_model}

\paragraph{Training.}
We start from the released LLaDA2.1-mini checkpoint and train LoRA adapters ($r{=}16$, $\alpha{=}32$, Kaiming initialization) on the query-key-value and dense projection modules.
The training set is a 50k-sample subset of FineWeb-Edu (score $\ge 3$, 256--1800 tokens per document), formatted as standard continued-pretraining prose with no task-specific labels.
We chose this corpus for convenience rather than through a systematic data study: it provides clean English expository text at modest scale, but we did not optimize mixture design, domain coverage, or error-type targeting.
It is unrelated to the downstream evaluation benchmarks in \cref{sec:experiments}, and stronger results may require more deliberate dataset engineering---for example, curating text with numeric revisions, structured reasoning, or other edit-rich patterns aligned with downstream failure modes.
Each example uses \ours{} with two forward passes per step: an M2T mask ratio sampled uniformly from $[0.3,0.8]$, a T2T edit ratio from $[0.1,0.3]$, and a 5\% random-token mixture on back-filled positions; draft remask ratios are sampled from $[0.0,0.3]$.
We optimize with AdamW at learning rate $5{\times}10^{-6}$ for one epoch (global batch size 8, sequence length 2048, cosine schedule with 1\% warmup) under FSDP2 with bf16 and gradient checkpointing, using loss weights $(\lambda_{\mathrm{m2t}},\lambda_{\mathrm{edit}},\lambda_{\mathrm{clean}})=(1.0,0.3,0.2)$.
We evaluate the merged adapter at training step 6250.

\paragraph{Inference.}
All benchmark numbers use the unchanged LLaDA2.1 official original-mode T2T inference procedure and parameters: the released \texttt{generate()} path with Q-Mode thresholds from the model card ($\tau_{\mathrm{m2t}}{=}0.7$, $\tau_{\mathrm{t2t}}{=}0.5$, \texttt{block\_length}${=}32$, \texttt{max\_post\_steps}${=}16$, temperature ${=}0$, greedy decoding).
We do not use remasking, token-to-mask decoding, multi-block editing, or per-benchmark threshold tuning.
Base and \ours{} checkpoints share the same inference code and the same official hyperparameters; only the weights differ.
For AIME~2025 we use the boxed-prompt evaluator consistent with prior LLaDA2.1 reporting.
We report task accuracy and four trace-level diagnostics: average output tokens, average T2T edits, edits per 100 output tokens (E/100tok), and average forward passes.

\section{Experiments and Results}
\label{sec:experiments}

\subsection{Evaluation Protocol}

We compare the released LLaDA2.1-mini base model against our FineWeb CPT LoRA adapter trained with \ours{} under the shared protocol in \cref{sec:data_model}.
Table~\ref{tab:main_results} reports accuracy and generation traces on CMATH, TriviaQA, PIQA, and AIME~2025.

\paragraph{Benchmarks.}
The reported set covers four answer regimes.
TriviaQA is an open-domain factual QA benchmark whose validation examples are scored by exact match against answer aliases~\citep{joshi2017triviaqa}.
PIQA is a two-choice physical commonsense benchmark, scored by option accuracy~\citep{bisk2020piqa}.
CMATH is a Chinese elementary-school math benchmark with short numeric final answers, scored by numeric exact match~\citep{wei2023cmath}.
AIME~2025 is a competition-math benchmark with integer answers; we report pass@1 on the 30-problem 2025 set.
Table~\ref{tab:benchmark_details} gives the split, size, answer format, and metric used in our runs.

\begin{table}[t]
\centering
\caption{Evaluation benchmarks used in our runs.}
\label{tab:benchmark_details}
\small
\begin{tabular}{lllll}
\toprule
Benchmark & Split & Samples & Answer format & Metric \\
\midrule
TriviaQA & validation & 17,944 & entity or short phrase, with aliases & EM \\
PIQA & validation & 1,838 & two-way choice, A/B or option text & Accuracy \\
CMATH & test & 1,098 & final numeric answer & Numeric EM \\
AIME~2025 & test & 30 & integer answer & Pass@1 \\
\bottomrule
\end{tabular}
\end{table}

\subsection{Main Results}

\begin{table}[t]
\centering
\caption{Results under official Q-Mode T2T inference (\cref{sec:data_model}). $\Delta$ columns report \oursshort{} minus Base. Higher is better for accuracy; lower for tokens, edits, E/100tok, and forward passes.}
\label{tab:main_results}
\scriptsize
\resizebox{\textwidth}{!}{%
\begin{tabular}{lrrrrrrrrrrrrrrr}
\toprule
\multirow{2}{*}{Benchmark}
& \multicolumn{3}{c}{Accuracy (\%)}
& \multicolumn{3}{c}{Avg.\ output tokens}
& \multicolumn{3}{c}{Avg.\ T2T edits}
& \multicolumn{3}{c}{E/100tok}
& \multicolumn{3}{c}{Avg.\ forward passes} \\
\cmidrule(lr){2-4}\cmidrule(lr){5-7}\cmidrule(lr){8-10}\cmidrule(lr){11-13}\cmidrule(lr){14-16}
& Base & \oursshort{} & $\Delta$
& Base & \oursshort{} & $\Delta$
& Base & \oursshort{} & $\Delta$
& Base & \oursshort{} & $\Delta$
& Base & \oursshort{} & $\Delta$ \\
\midrule
TriviaQA   & 43.71 & 44.59 & +0.88 & 113.30 & 92.87  & $-$20.43 & 5.11  & 2.39  & $-$2.72 & 4.51 & 2.57 & $-$1.94 & 66.10  & 63.69  & $-$2.41 \\
PIQA       & 82.37 & 84.00 & +1.63 & 112.88 & 92.18  & $-$20.70 & 3.04  & 1.64  & $-$1.40 & 2.69 & 1.78 & $-$0.91 & 61.67  & 57.34  & $-$4.33 \\
CMATH      & 82.33 & 87.80 & +5.47 & 279.94 & 435.36 & +155.42 & 11.92 & 7.54  & $-$4.38 & 4.26 & 1.73 & $-$2.53 & 101.14 & 255.14 & +154.00 \\
AIME~2025  & 30.00 & 30.00 & 0.00  & 5584.87 & 5898.40 & +313.53 & 130.20 & 86.00 & $-$44.20 & 2.33 & 1.46 & $-$0.87 & 1548.47 & 1565.80 & +17.33 \\
\bottomrule
\end{tabular}}
\end{table}

\paragraph{Interpretation.}
The improvements are modest on TriviaQA and PIQA but large on CMATH, and the trace statistics show that the checkpoint does not uniformly shorten outputs.
\ours{} lowers edit intensity (E/100tok) on all four benchmarks.
Accuracy improves most where flip analysis shows the base model already reasons correctly but commits the wrong final token (CMATH), or where the base model over-edits long factual outputs before giving a short answer (TriviaQA and PIQA).
On TriviaQA and PIQA, \ours{} reduces average output length and T2T edits while improving accuracy.
This pattern is consistent with fewer destructive open-ended self-correction loops rather than learning new task-specific answer formats.
On CMATH, accuracy rises by 5.47 points even though average output length \emph{increases} from 279.94 to 435.36 tokens.
Flip analysis shows that most CMATH gains come from fixing final numeric commitment errors---for example, writing \texttt{4440} instead of \texttt{1440} after deriving the correct intermediate result---while T2T edits decrease from 11.92 to 7.54.
Because the CPT corpus contains no Chinese math problems, this gain is better explained by mechanism transfer from editing plausible numeric draft errors in English prose than by memorizing CMATH-style templates.
Flip analysis also shows that \ours{} still makes digit-commit errors on 52/1097 CMATH items (multi-digit, missing-digit, or scale errors), so the mechanism reduces but does not eliminate this failure mode.
AIME~2025 matches the base score at 9/30 under the boxed prompt; average edits drop from 130.2 to 86.0 even though outputs remain long ($\sim$5.6k tokens).

\subsection{Qualitative Cases}
\label{sec:qualitative}

We highlight three flip examples.
Each box reports the question, gold answer, truncated model outputs, and trace statistics.

\casebox{%
\small
\textbf{Case 1 (CMATH): final-digit commitment.}\\
\textbf{Question:} A bus travels 480\,km in 8 hours. At the same speed, how far apart are two cities if the travel time is 24 hours?\\
\textbf{Gold:} 1440\\
\textbf{Base} (164 tok, 17 edits): \emph{First, find the speed: $480/8=60$\,km/h. Then $60\times24=1440$\,km \ldots\ [middle steps omitted] \ldots\ therefore the distance is \textbf{4440}\,km.}\\
\textbf{\oursshort{}} (132 tok, 11 edits): \emph{Speed is $480/8=60$\,km/h, so $60\times24=1440$\,km \ldots\ the answer is \textbf{1440}.}\\
\textbf{Takeaway:} Both models derive 1440, but the base model corrupts the final numeric token; \oursshort{} commits the correct digit with fewer edits.}

\casebox{%
\small
\textbf{Case 2 (CMATH): trailing-digit error.}\\
\textbf{Question:} A factory saves 39 tons of coal each day. How many tons are saved in 30 days?\\
\textbf{Gold:} 1170\\
\textbf{Base:} \emph{Multiply: $39\times30=1170$ \ldots\ [explanation continues] \ldots\ the total saving is \textbf{11770}\,tons.}\\
\textbf{\oursshort{}} (1 edit): \emph{$39\times30=1170$; answer \textbf{1170}.}\\
\textbf{Takeaway:} The base model keeps the correct intermediate product but appends an extra digit at the final commitment step.}

\casebox{%
\small
\textbf{Case 3 (TriviaQA): long self-correction loop.}\\
\textbf{Question:} Areas on the Canadian eastern coast share their name with two breeds of dog, Newfoundland and which other?\\
\textbf{Gold:} Labrador\\
\textbf{Base} (7819 tok, 59 edits): \emph{The NHL team originally named after a Disney movie is the Tampa Lightning \ldots\ also mentioned are the Anaheim Ducks \ldots\ [thousands of tokens of repeated rewrites and negations omitted] \ldots\ still no extractable correct breed.}\\
\textbf{\oursshort{}} (49 tok, 0 edits): \emph{labrador}\\
\textbf{Takeaway:} The base model enters a long open-ended revision loop; \oursshort{} stops early with a short, scorable answer.}

These cases match the two patterns discussed above: CMATH gains come from fixing final numeric commitment errors, while TriviaQA gains mainly reflect shorter outputs with less destructive editing.

\section{Discussion and Conclusion}
\label{sec:discussion}

This work should be read as a first training-side study of T2T distribution alignment, not as a complete benchmark win.
The evidence supports three points.
First, the mismatch between random-token T2T training and model-generated T2T inference is concrete enough to motivate a different training objective.
Second, self-generated T2T corruptions can change the behavior of the native LLaDA2.1 editor without modifying inference, as shown by CMATH gains without any Chinese math in the CPT corpus.
Third, the benefit is error-type dependent: the method helps most when the failure is a bad final commitment or excessive revision, not when the model lacks the underlying reasoning needed to solve the task.

The next version of the method should include a full ablation grid over edit loss weight, random-token ratio, remask ratio, data composition, and LoRA rank, plus a head-to-head random-token T2T baseline trained on the same data and compute budget.
Future work should also study dataset design more carefully, rather than relying on a convenient generic CPT corpus, and validate the mechanism on additional datasets and task formats beyond those studied here.
The current result nevertheless identifies a clear research direction: instead of only changing how diffusion LMs revise tokens at inference time, we can train their native editors on the structured mistakes they actually make.

\bibliographystyle{plainnat}
\bibliography{refs}

@inproceedings{austin2021d3pm,
  title={Structured Denoising Diffusion Models in Discrete State-Spaces},
  author={Austin, Jacob and Johnson, Daniel D. and Ho, Jonathan and Tarlow, Daniel and van den Berg, Rianne},
  booktitle={Advances in Neural Information Processing Systems},
  year={2021}
}

@inproceedings{lou2024sedd,
  title={Discrete Diffusion Modeling by Estimating the Ratios of the Data Distribution},
  author={Lou, Aaron and Meng, Chenlin and Ermon, Stefano},
  booktitle={International Conference on Machine Learning},
  year={2024}
}

@inproceedings{sahoo2024mdlm,
  title={Simple and Effective Masked Diffusion Language Models},
  author={Sahoo, Subham Sekhar and Arriola, Marianne and Schiff, Yair and Gokaslan, Aaron and Marroquin, Edgar and Chiu, Justin T. and Rush, Alexander M. and Kuleshov, Volodymyr},
  booktitle={Advances in Neural Information Processing Systems},
  year={2024}
}

@inproceedings{shi2024md4,
  title={Simplified and Generalized Masked Diffusion for Discrete Data},
  author={Shi, Jiaxin and Han, Kehang and Wang, Zhe and Doucet, Arnaud and Titsias, Michalis K.},
  booktitle={Advances in Neural Information Processing Systems},
  year={2024}
}

@inproceedings{ou2025radd,
  title={Your Absorbing Discrete Diffusion Secretly Models the Conditional Distributions of Clean Data},
  author={Ou, Jingyang and Nie, Shen and Xue, Kaiwen and Zhu, Fengqi and Sun, Jiacheng and Li, Zhenguo and Li, Chongxuan},
  booktitle={International Conference on Learning Representations},
  year={2025}
}

@article{nie2025llada,
  title={Large Language Diffusion Models},
  author={Nie, Shen and Zhu, Fengqi and You, Zebin and Zhang, Xiaolu and Ou, Jingyang and Hu, Jun and Zhou, Jun and Lin, Yankai and Wen, Ji-Rong and Li, Chongxuan},
  journal={arXiv preprint arXiv:2502.09992},
  year={2025}
}

@article{dream7b,
  title={Dream 7B: Diffusion Large Language Models},
  author={Ye, Jiacheng and Xie, Zhihui and Zheng, Lin and Gao, Jiahui and Wu, Zirui and Jiang, Xin and Li, Zhenguo and Kong, Lingpeng},
  journal={arXiv preprint arXiv:2508.15487},
  year={2025}
}

@article{gong2025diffullama,
  title={Scaling Diffusion Language Models via Adaptation from Autoregressive Models},
  author={Gong, Shansan and Agarwal, Shivam and Zhang, Yizhe and Ye, Jiacheng and Zheng, Lin and Li, Mukai and An, Chenxin and Zhao, Peilin and Bi, Wei and Han, Jiawei and Peng, Hao and Kong, Lingpeng},
  journal={International Conference on Learning Representations},
  year={2025}
}

@article{inception2025mercury,
  title={Mercury: Ultra-Fast Language Models Based on Diffusion},
  author={{Inception Labs} and Khanna, Samar and Kharbanda, Siddhant and Li, Shufan and Varma, Harshit and Wang, Eric and Birnbaum, Sawyer and Luo, Ziyang and Miraoui, Yanis and Palrecha, Akash and Ermon, Stefano and Grover, Aditya and Kuleshov, Volodymyr},
  journal={arXiv preprint arXiv:2506.17298},
  year={2025}
}

@inproceedings{arriola2025bd3lm,
  title={Block Discrete Denoising Diffusion Language Models},
  author={Arriola, Marianne and Gokaslan, Aaron and Chiu, Justin T. and Yang, Zhihan and Qi, Zhixuan and Han, Jiaqi and Sahoo, Subham Sekhar and Kuleshov, Volodymyr},
  booktitle={International Conference on Learning Representations},
  year={2025}
}

@article{kang2025parallelbench,
  title={{ParallelBench}: Understanding the Trade-offs of Parallel Decoding in Diffusion {LLMs}},
  author={Kang, Wonjun and Galim, Kevin and Oh, Seunghyuk and Lee, Minjae and Zeng, Yuchen and Zhang, Shuibai and Hooper, Coleman and Hu, Yuezhou and Koo, Hyung Il and Cho, Nam Ik and Lee, Kangwook},
  journal={arXiv preprint arXiv:2510.04767},
  year={2025}
}

@article{llada21,
  title={LLaDA2.1: Speeding Up Text Diffusion via Token Editing},
  author={Bie, Tiwei and Cao, Maosong and Cao, Xiang and Chen, Bingsen and Chen, Fuyuan and Chen, Kun and Du, Lun and Feng, Daozhuo and Feng, Haibo and Gong, Mingliang and others},
  journal={arXiv preprint arXiv:2602.08676},
  year={2026}
}

@article{wang2025remdm,
  title={Remasking Discrete Diffusion Models with Inference-Time Scaling},
  author={Wang, Guanghan and Schiff, Yair and Sahoo, Subham Sekhar and Kuleshov, Volodymyr},
  journal={arXiv preprint arXiv:2503.00307},
  year={2025}
}

@inproceedings{huang2026remedi,
  title={Don't Settle Too Early: Self-Reflective Remasking for Diffusion Language Models},
  author={Huang, Zemin and Wang, Yuhang and Chen, Zhiyang and Qi, Guo-Jun},
  booktitle={International Conference on Learning Representations},
  year={2026}
}

@article{remask_dont_replace,
  title={Remask, Don't Replace: Token-to-Mask Refinement in Diffusion Large Language Models},
  author={Yao, Lin},
  journal={arXiv preprint arXiv:2604.18738},
  year={2026}
}

@article{schiff2026proseco,
  title={Learn from Your Mistakes: Self-Correcting Masked Diffusion Models},
  author={Schiff, Yair and Belhasin, Omer and Uziel, Roy and Wang, Guanghan and Arriola, Marianne and Turok, Gilad and Elad, Michael and Kuleshov, Volodymyr},
  journal={arXiv preprint arXiv:2602.11590},
  year={2026}
}

@article{liu2026backplay,
  title={BackPlay: Head-Only Look-Back Self-Correction for Diffusion Language Models},
  author={Liu, Liming and Huang, Binxuan and Zhang, Zixuan and Liu, Xin and Yin, Bing and Zhao, Tuo},
  journal={arXiv preprint arXiv:2601.06428},
  year={2026}
}

@inproceedings{bengio2015scheduled,
  title={Scheduled Sampling for Sequence Prediction with Recurrent Neural Networks},
  author={Bengio, Samy and Vinyals, Oriol and Jaitly, Navdeep and Shazeer, Noam},
  booktitle={Advances in Neural Information Processing Systems},
  year={2015}
}

@inproceedings{ross2011dagger,
  title={A Reduction of Imitation Learning and Structured Prediction to No-Regret Online Learning},
  author={Ross, St{\'e}phane and Gordon, Geoffrey J. and Bagnell, J. Andrew},
  booktitle={International Conference on Artificial Intelligence and Statistics},
  year={2011}
}

@inproceedings{chen2023analogbits,
  title={Analog Bits: Generating Discrete Data using Diffusion Models with Self-Conditioning},
  author={Chen, Ting and Zhang, Ruixiang and Hinton, Geoffrey},
  booktitle={International Conference on Learning Representations},
  year={2023}
}

@article{joshi2017triviaqa,
  title={{TriviaQA}: A Large Scale Distantly Supervised Challenge Dataset for Reading Comprehension},
  author={Joshi, Mandar and Choi, Eunsol and Weld, Daniel S. and Zettlemoyer, Luke},
  journal={arXiv preprint arXiv:1705.03551},
  year={2017}
}

@article{bisk2020piqa,
  title={{PIQA}: Reasoning about Physical Intuition in Natural Language},
  author={Bisk, Yonatan and Zellers, Rowan and Bras, Ronan Le and Gao, Jianfeng and Choi, Yejin},
  journal={arXiv preprint arXiv:1911.11641},
  year={2020}
}

@article{wei2023cmath,
  title={{CMATH}: Can Your Language Model Pass Chinese Elementary School Math Test?},
  author={Wei, Tianwen and Luan, Jian and Liu, Wei and Dong, Shuang and Wang, Bin},
  journal={arXiv preprint arXiv:2306.16636},
  year={2023}
}

\end{document}